\documentclass[letterpaper, 10 pt, conference]{ieeeconf} 

\IEEEoverridecommandlockouts                             

\usepackage{graphics}
\usepackage{epsfig}
\usepackage{amsmath} 
\usepackage{multirow}
\usepackage{float}
\usepackage{booktabs}
\usepackage{url}
\usepackage{algorithmic}
\usepackage{algorithm}
\newcommand{\etal}{\textit{et al}.}

\newcommand{\argmin}{\mathop{\rm argmin}\limits}

\title{\LARGE \bf
    Neural Implicit Event Generator for Motion Tracking
}

\author{Mana Masuda$^{1*}$, Yusuke Sekikawa$^{2*}$, Ryo Fujii$^{1}$, Hideo Saito$^{1}$
\thanks{* Equal contribution}
\thanks{$^{1}$Mana Masuda, Ryo Fujii and Hideo Saito are with the School of Science for Open and Environmental Systems, Keio University, Tokyo, Japan}%
\thanks{$^{2}$Yusuke Sekikawa is with Denso IT Laboratory, Tokyo, Japan}%
}

\usepackage{capt-of,etoolbox}
\begin{document}

\makeatletter
\apptocmd\@maketitle{{\myfigure{}\par}}{}{}

\newcommand\myfigure{%
\centering
\includegraphics[width=0.95\hsize]{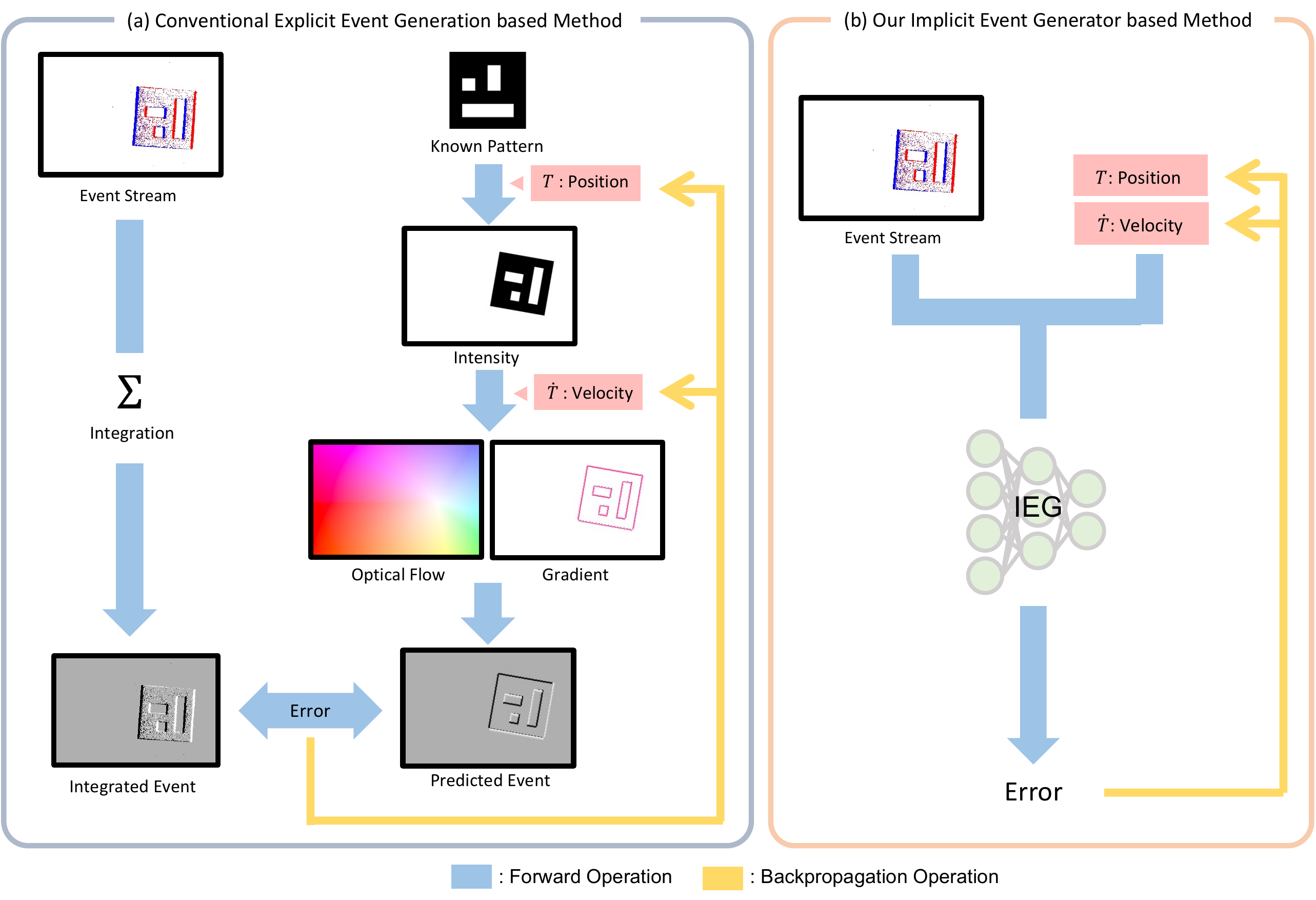}
\captionof{figure}{(a) The conventional explicit event generation based tracking methods. (b) Our proposed implicit event generator (IEG) based tracking method. The existing methods require to generate events explicitly with dense computation for computing error to update estimation of the position $T$ and velocity $\Dot{T}$.
Ours can estimate the $T$ and $\dot{T}$ with sparse computation by using IEG which implicitly model event generation process. By utsing our framework using IEG, error is computed directly with sparse computation without explicitly computing event frame from $T$ and $\Dot{T}$ using dense computation}
\label{fig:caricature}
}

\maketitle
\thispagestyle{empty}
\pagestyle{empty}
\setcounter{figure}{1}

\begin{abstract}
We present a novel framework of motion tracking from event data using implicit expression. Our framework use pre-trained event generation MLP named \textit{implicit event generator} (IEG) and does motion tracking by updating its state (position and velocity) based on the difference between the observed event and generated event from the current state estimate.
The difference is computed implicitly by the IEG. 
Unlike the conventional explicit approach, which requires dense computation to evaluate the difference, our implicit approach realizes efficient state update directly from sparse event data. 
Our sparse algorithm is especially suitable for mobile robotics applications where computational resources and battery life are limited.
To verify the effectiveness of our method on real-world data, we applied it to the AR marker tracking application.
We have confirmed that our framework works well in real-world environments in the presence of noise and background clutter.
\end{abstract}

\section{INTRODUCTION}
\begin{figure*}[htb]
    \centering
    \includegraphics[width=0.9\hsize]{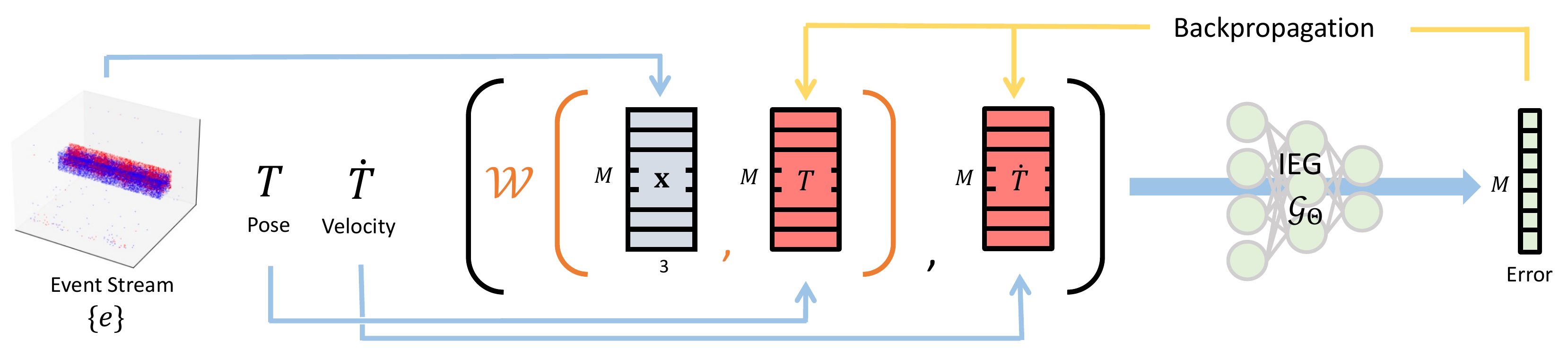}
    \caption{An overview of our framework of object tracking pipeline which inverts an optimized implicit event generator (IEG). Given an initially estimated position $T$ and velocity $\Dot{T}$ and the place and time where intensity change is detected, IEG estimate the corresponding intensity change and gives its error. Since the whole pipeline is differentiable, we can refine our estimated position $T$ and velocity $\Dot{T}$ by minimizing the error. All operations are done sparsely only where the events are detected.}
    \label{fig:overview}
\end{figure*}

Motion tracking from spatiotemporal data such as video frames is one of the fundamental functionality for robotics applications. 
It either tracks a specific target position from the camera or tracks a camera position from worlds coordinates.
The application of tracking algorithms ranges from robotic arm manipulation, and mobile robot localization.
Before the tracking, the target's initial position is identified by globally matching known patterns with camera observations.
And then, a tracking algorithm sequentially updates the position using. 
This update is usually done by using gradient-based algorithms. 
However, in existing systems using frame cameras \cite{mur2015orb,forster2016svo}, the loss function defined by the difference between the known pattern and the observation is often trapped by the local minimum when the motion of the camera or the target is fast; results in the failure of tracking.

Event cameras are bio-inspired novel vision sensors that mimic biological retinas and report per-pixel intensity changes in the form of asynchronous event stream instead of intensity frames. 
Due to the unique operating principle, event-based sensing offers significant advantages over conventional cameras; high dynamic range (HDR), high temporal resolution, and blurless sensing. 
As a result, it has the potential to achieve robust tracking in intense motion and severe lighting environments.

Many methods have been proposed which use these features to achieve tracking in high-speed or high-dynamic-range environments \cite{rebecq2017real, zihao2017event, gallego2017event, gehrig2018asynchronous, bryner2019event, alzugaray2020haste, chamorro2020high, gehrig2020eklt}. 
In particular, Bryner \etal~\cite{bryner2019event} and Gehrig \etal~\cite{gehrig2018asynchronous, gehrig2020eklt} proposed method 
as an extension of Lucas-Kanade tracker (KLT) \cite{10.5555/1623264.1623280, Baker2004LucasKanade2Y} developed for frame-based video data.
They extend KLT to sparse intensity difference data from the event-based camera. 
In their method, position and velocity are updated by minimizing the difference between the observation and the expected observation, as shown in Fig. \ref{fig:caricature}(a).
The expected observation is computed by warping the given photometric 2D/3D map using the current estimate of position and velocity.
Unfortunately, these methods could not fully take advantage of the sparseness of event data. 
Still, they require dense computation to compute the difference, and this optimization needs to be repeated many times until convergence.

To realize efficient tracking, we focus on leveraging the sparsity of event data. 
To this end, we propose a novel framework for object tracking using implicit expression. 
As shown in Fig. \ref{fig:caricature}(b), our method achieves the goal by updating the position and velocity based on the difference between the events generated by the implicit event generator (IEG) and the observed events. 
In our method, the difference is directly computed by the feed-forward operation of sparse event data without explicitly predict and generate events from the current position and velocity estimates.
This implicit approach is totally different from the conventional explicit approach, which requires dense computation to evaluate the difference. 
The gradient of position and velocity w.r.t the IEG is computed by simply backpropagating the error through IEG realized by MLP. 
The entire operation, namely feed-forward operation for difference evaluation and derivative computation w.r.t the difference, are executed sparsely; computation happens only where the events occur. 

In this paper, we first introduce a specific tracking algorithm based on this new framework. For the sake of proof of concept, we conducted experiments using an AR marker captured by an actual event-based camera to show that the method based on this new framework can be applied to real problems.

Our main contributions are summarized as follows:
\begin{itemize}
    \item We propose a novel object tracking framework by using an implicit event generator (IEG). In our method, computation is performed only at the event's location; therefore, it could be significantly efficient than the conventional explicit method requiring dense computation.
    \item We conduct experiments using data captured from AR markers using a real event-based camera; demonstrate the proposed method can be applied to practical problems.
\end{itemize}

\section{RELATED WORKS}
Most of the existing tracking algorithms are developed based on RGB cameras and track the target object frame by frame. Recently, the Siamese network-based approach achieve a better result and many methods have been proposed based on it \cite{bertinetto2016fully, li2018high, voigtlaender2020siam}. RGB images, however, cannot provide sufficient tracking performance under fast motion and low-illumination conditions.

\begin{figure*}[tb]
    \centering
    \includegraphics[width=1.0\hsize]{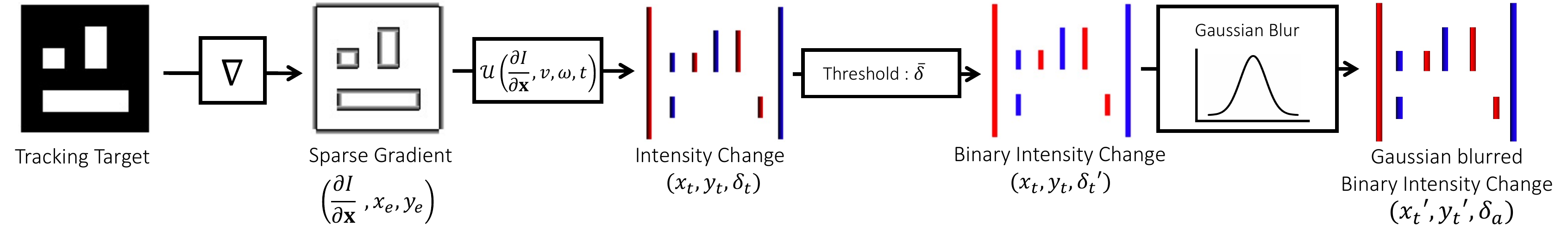}
    \caption{Procedure for creating artificial intensity change data for training IEG. 
The detailed process is as follows: 1) compute the gradient of tracking target intensity image; 2) randomly determine the time $t$ and velocity $v$ and the rotation velocity $\omega$, and calculate the intensity change value $\delta_t$ when moving at that velocity and time; 3) based on the threshold value $\bar{\delta}$, determine whether the intensity change value $\delta_t$  exceeds the change detection threshold of an event-based camera to generate binary intensity changes where  $\delta_t' = (+1, -1, \mathtt{Nan})$; 4) we apply Gaussian blur $G(x, y)$ to the calculated intensity change $\delta_t'$ to obtain the artificial intensity change data $\delta_a$ to train IEG.}
    \label{fig:datamaking_scheme}
\end{figure*}

By taking advantage of the HDR and no-motion blur properties of event cameras, a method for tracking fast objects and in harsh light environments has been proposed. Rebecq \etal~\cite{rebecq2017real} and Zihao \etal~\cite{zihao2017event} convert IMU data into lower rate pieces of information at desired times where events are collected. Chamorro \etal~\cite{chamorro2020high} and Alzugaray \etal~\cite{alzugaray2020haste} explore the high-speed feature tracking with the dynamic vision sensor (DVS) \cite{4444573}, and Gallego \etal~\cite{gallego2017event} presented a filter-based approach that requires to maintain a history of past camera positions which are continuously refined. Bryner \etal~\cite{bryner2019event} and Gehrig \etal~\cite{gehrig2018asynchronous, gehrig2020eklt} use the extended Lucas-Kanade method \cite{10.5555/1623264.1623280, Baker2004LucasKanade2Y} and update the position and velocity by warping the given photometric 3D map or frame and taking the difference between the observation and the expected observation.

Considering the processing in the neural network, sparse event data is inherently difficult to handle using a convention deep neural network.
Therefore, many neural network-based methods for event data first convert sparse events to dense frames and then process it by a dense neural network such as CNN \cite{maqueda2018event, baldwin2021time, tulyakov2019learning, gehrig2019end, cannici2020matrix}.
Comparing with these methods, the computational complexity of our approach could be significantly smaller.	

Since NeRF \cite{mildenhall2020nerf}, implicit volumetric representations have attracted in computer vision community \cite{mescheder2019occupancy, park2020deformable, martin2021nerf}. In this paradigm, the RGB-D scene representation is learned in a parameterized manner by a multi-layer perceptron (MLP). Some methods have been proposed to estimate the position of an object by back-propagating the scene representation MLP, such as iNeRF \cite{yen2020inerf, su2021nerf}. These methods, however, are proposed for dense RGB images and do not apply to sparse event data with time-series information. We used IEG to represent events that occur in response to the position, time, and velocity of a pixel, making it possible to implicitly represent events under a variety of conditions.

\section{Method}
We present a framework for motion tracking using a neural implicit event generator (IEG)  to realize efficient motion update by sparse operations.
In our framework, position $T$ and velocity $\Dot{T}$ is updated by simply feed-forwarding sparse event observation into IEG and then backpropagate the output through the IEG.
The IEG $\mathcal{G}$ is realized by differential MLP (parameterized by $\Theta$), and it is trained before inference to output intensity changes of tracking-target for given position $T$ and velocity $\Dot{T}$.
To track the motion, i.e., estimate $(T, \Dot{T})$, at inference, $(T, \Dot{T})$ is otpmized to minimized the difference $\mathcal{L}$ betweeen obserbation and estimation for observation of event stream $\{e\}$ such that:
\begin{equation}
    \label{eq: gradient_optimization_definition}
   \hat{T}, \hat{\Dot{T}} = \argmin_{T, \Dot{T}} \mathcal{L}(T, \Dot{T}|\{e\}, \Theta) 
\end{equation}
In this formulation,  explicit generation of events using by current estimate $(T, \Dot{T})$ does not required as existing method (Fig.\ref{fig:caricature}, left). 
Instead, ours implicitly generates events and computes the difference directly from the sparse event (Fig.\ref{fig:caricature}, right).

This paper applied the proposed framework for $3$DoF camera motion in a $2$D planer scene.

\subsection{Formulation of IEG}
We represent intensity change appearance as a $6$D vector-valued function whose input is a $2$D location $\mathbf{x} = (x, y, r)$ and time $t$ and $3$DoF $\Dot{T} = (v_{x}, v_{y}, \omega)$, and whose output is an detected intensity change $\delta \in [+/-/\texttt{NaN}]$. We approximate the intensity change appearance with continuous $3$DoF with the IEG $\mathcal{G}_{\Theta}:(x,y,t,v_{x},v_{y},\omega)\rightarrow \delta$ and optimize its weights $\Theta$ to map from each input $6$D vector to its corresponding intensity change $\delta$.

\subsection{Artificial Data of Intensity Changement}
\label{seq:artificial-data}
To learn the intensity change with velocity using IEG, we create artificial intensity change data based on the following equation; 
\begin{equation}
    \left|\frac{\partial I}{\partial {\mathbf{x}}} \cdot \mathbf{v(\mathbf{x})} \Delta t \right| > C,
\end{equation}
where $\mathbf{v}(\mathbf{x})$ is the speed of coordinate $\mathbf{x}$. $C$ is the threshold of event firing.

To create intensity change data at various velocities, we use a randomly determined velocity $\Dot{T} = ({v}_x, {v}_y, \omega)$, and from the edge coordinates of a known object $(x_e, y_e)$ and the intensity derivative data at those coordinates $\frac{\partial I}{\partial \mathbf{x}}=(\partial i_x, \partial i_y)$. From these values, we can calculate the theoretical intensity change value $\delta_{t}$ at the coordinate $(x_t, y_t)$, which corresponding to the edge coordinate $(x_e, y_e)$ when the object moves by time $t$ as follows;
\begin{equation}
\begin{split}
    & \delta_t = \mathcal{U}\left(\frac{\partial I}{\partial \mathbf{x}}, v, \omega, t \right)\\
             &= \begin{pmatrix}
                \cos(\omega t) & -\sin(\omega t) \\
                \sin(\omega t) & \cos(\omega t)\\
               \end{pmatrix}
              \begin{pmatrix}
                 \partial i_x \\
                 \partial i_y \\
              \end{pmatrix}
               \cdot
               \begin{pmatrix}
                v_x + \cos(\omega) \\
                v_y + \sin(\omega) \\
               \end{pmatrix}
               t
\end{split}
\end{equation}
We set the value corresponding to the threshold of event firing as $\bar{\delta}$, and define $\delta_t'$ as follows;
\begin{equation}
    \delta_t' = 
    \left\{
    \begin{array}{ll}
    +1 & (\delta_{t} > \bar{\delta}) \\
    \texttt{Nan} & (-\bar{\delta} <= \delta_{t} <= \bar{\delta})\\
    -1 & (\delta_{t} < -\bar{\delta})
    \end{array}.
    \right.
\end{equation}

To perform the optimization operation described in \ref{sec:Method-Optimization}, we applied a Gaussian blur to the intensity change $\delta_t'$ to make it differentiable. Depending on the distance $(dx, dy)$ to the intensity change $\delta_t'$, we use an extended Gaussian function $G(dx, dy)$ with mean $0$ and variance $\sigma$;
\begin{equation}
    G(dx, dy) = \exp{\left(-\frac{{dx}^2+{dy}^2}{2 \sigma^2} \right)}.
\end{equation}
To avoid interference with events generated by other edges, the blur width is set to $w$, and the Gaussian blurred intensity change $\delta_a$ at $(x_t', y_t')=(x_t+dx, y_t+dy)$ is calculated as follows:
\begin{equation}
    \delta_a = 
    \left \{
    \begin{array}{ll}
    G(dx, dy) \times \delta_t' &(\sqrt{dx^2+dy^2}\leq w)\\
    0                          &(\sqrt{dx^2+dy^2}>w)
    \end{array}.
    \right.
\end{equation}

\subsection{Training IEG}
We trained the IEG with the artificial data as described in \ref{seq:artificial-data}. When we input the $6$D vector of artificial data $(x, y, t, {v}_x, {v}_y, \omega)$ and the IEG outputs the estimated intensity change $\delta_p$. The IEG is updated by taking the difference between the generated intensity change $\delta_g$ and the artificially calculated intensity change $\delta_a$;
\begin{equation}
    \mathcal{L}_{train} = ||\delta_g - \delta_a||_2.
\end{equation}

\begin{algorithm}[tb]
\label{algo:gradient-based optimization}
    \caption{Tracking using gradient of IEG}
    \label{alg1}
    \begin{algorithmic}[1]
    \ENSURE $\hat{T}_k = (x_k, y_k)$ is the inferenced position at $k$-th iteration and $\hat{\dot{T}}_k = ({v_x}_k, {v_y}_k, \omega_k)$ is the inferenced velocity at $k$-th iteration.
    \STATE $\epsilon \leftarrow inf$, $k \leftarrow 0$
    \STATE $\hat{T} \leftarrow T_0$
    \STATE $\hat{\dot{T}} \leftarrow (0, 0, 0)$
    \WHILE{$\epsilon \geq \bar{\epsilon}$}
    \STATE $\mathcal{L}_k = \sum_{i \in M}{(1-|\mathcal{G}_{\Theta}(\mathcal{W}(\mathbf{x}_i, \hat{T}_k), \hat{\dot{T}})|)}$
    \STATE Update $\hat{T}_k, \hat{\dot{T}}_k$
    \STATE $\varepsilon \leftarrow \mathcal{L}_{k-1} - \mathcal{L}_k$, $k \leftarrow k + 1$
    \ENDWHILE
    \end{algorithmic}
\end{algorithm}

\subsection{Tracking using gradient of IEG}
\label{sec:Method-Optimization}
Let $\Theta$ be the parameters of the trained and fixed IEG $\mathcal{G}$, $\mathbf{x}_i = (x_i, y_i, t_i)$ the place $(x_i, y_i)$ and time $t_i$ of $i$-th detected intensity change, $\delta(\mathbf{x}_i)$ is the intensity change sensed at $\mathbf{x}_i$, $(\hat{T}_k, \hat{\Dot{T}}_k)$ the estimated position and velocity at current optimization step $k$. Before inputting into the IEG, $\mathbf{x}_i$ is transformed by the transformation function $\mathcal{W}$ based on $T_k$ in the $2$D plane.
\begin{equation}
    \mathcal{W}(\mathbf{x}_i, T) = \begin{pmatrix}
                    \cos(-T_r) & -\sin(-T_r) \\
                    \sin(-T_r) & \cos(-T_r) \\
                  \end{pmatrix}
                  \begin{pmatrix}
                    x_i - {T_t}_x \\
                    y_i - {T_t}_y \\
                  \end{pmatrix}
,
\end{equation}
where $T_t$ is the translation component and $T_r$ is the rotation component of $T$. During tracking, optimization is performed so that the absolute value output by IEG is close to $1$;
\begin{equation}
    \mathcal{L}_{update} = \sum_{i \in M}{(1-|\mathcal{G}_{\Theta}(\mathcal{W}(\mathbf{x}_i, \hat{T}_k), \hat{\dot{T}})|)},
\end{equation}
where $M$ is the number of detected events. We employ gradient-based optimization to solve for $(\hat{T}, \hat{\Dot{T}})$ as defined in Eq. \ref{eq: gradient_optimization_definition}. The optimization is iterated until the change in loss $\epsilon$ is less than the optimization threshold $\bar{\epsilon}$.

In other words, this optimization problem can be defined as finding $(T, \dot{T})$ by solving the function $1-|\mathcal{G}|=0$.

\subsection{Object Tracking}
During tracking, the events of an existing small group (window) consisting of $M$ events are processed as in \ref{sec:Method-Optimization} to obtain the position $T_j$ and velocity $\Dot{T}_j$ at $j$-th iteration. Then, at the $j+1$-th iteration, slide the window with sliding width $K$, and update the position $T_{j+1}$ and velocity $\Dot{T}_{j+1}$ by optimizing again as in \ref{sec:Method-Optimization}. The position and velocity do not change much between the $j$-th and $j+1$-th iterations, so to update the $(T_j, \dot{T}_j)$ at $j+1$-th iteration could help to reduce the latency.

Our pipeline does not take acceleration $\ddot{T}$ into account, but this is not a problem because if the window size $M$ is sufficiently small, the motion within the window can be approximated as constant velocity motion.

\section{EXPERIMENTAL RESULTS} 
The significance of this method is that it can estimate the motion of an object only from the event data. To demonstrate the effectiveness of this method, we experimented using an image pattern with AR markers as a target scene to verify whether the motion of the AR marker can be accurately estimated or not. Since the events only occur from the edge of the AR marker, we can effectively show that the camera motion can be estimated using only sparse events. Another reason for using AR markers is that they are easy to annotate (detect and map feature points) and can be used as a criterion for accuracy evaluation. To capture the event data, we used Prophesee's event camera Gen3 \cite{prophesee} with $120$dB at $10000$fps. As the problem setup for object tracking, we can assume that the initial position of the object is known by the object detection algorithm, so we gave its approximate position as the initial value for tracking. The evaluation was done quantitatively and qualitatively. 

\begin{figure}[tb]
    \centering
    \includegraphics[width=0.8\hsize]{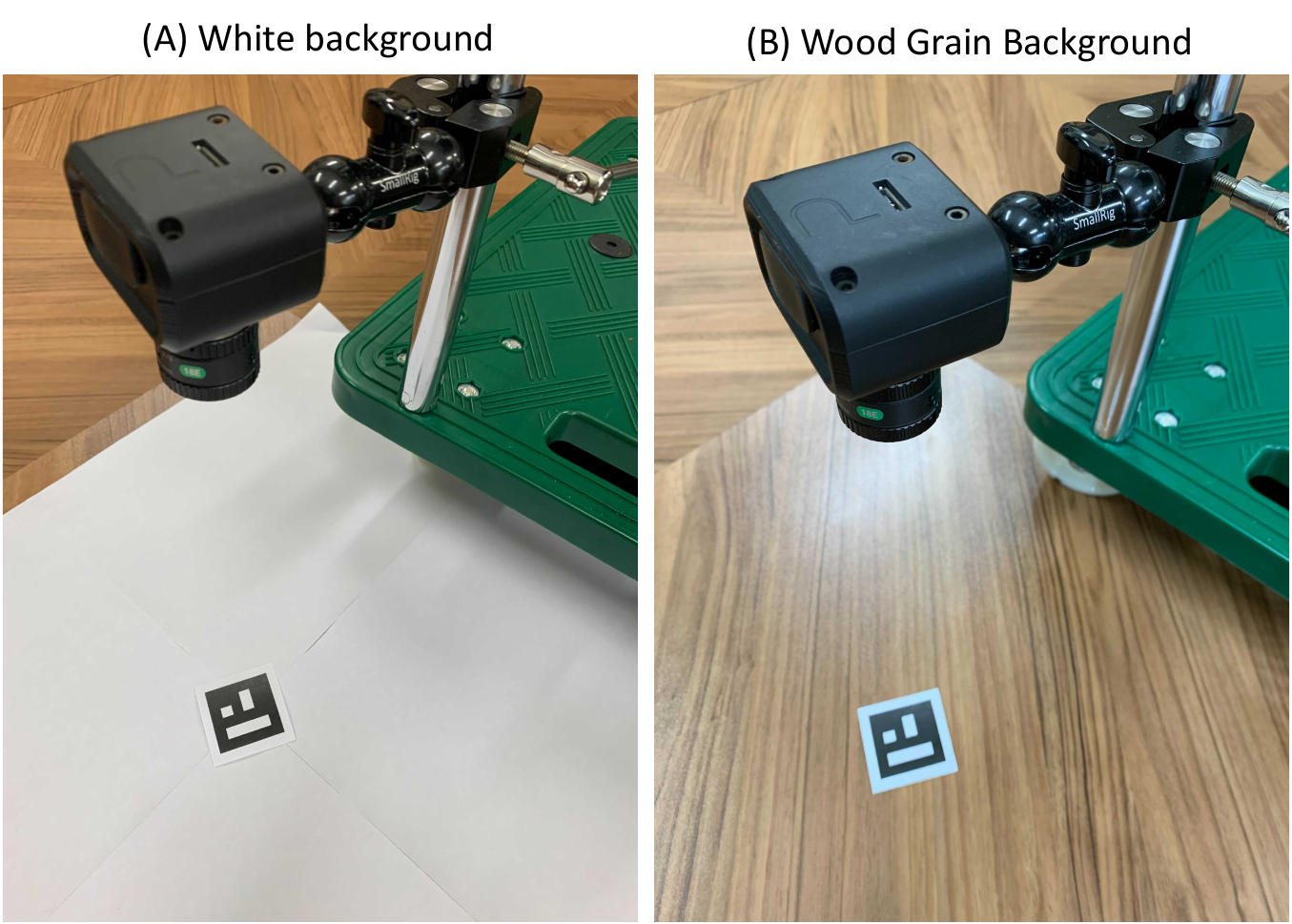}
    \caption{How to capture the data by using event camera \cite{prophesee}. (a) on a white background, and (b) on the wood grain pattern. Since our proposed tracking algorithm is limited to $2$ dimension, we fixed the camera on the dolly so that the distance between the camera and the AR marker is constant.}
    \label{fig:taking_event}
\end{figure}

\begin{table*}[tb]
    \centering
    \caption{Error measurements of on white background and on wood grain background}
    \label{tab:quantitative}
    \scalebox{0.72}[0.72]{
    \begin{tabular}{cccccccccccccccc}
    \cmidrule(lr){1-8} \cmidrule(lr){9-16}
                                \multicolumn{8}{c}{On White Background} &  \multicolumn{8}{c}{On Wood Grain Background} \\ \cmidrule(lr){1-8} \cmidrule(lr){9-16}
                          &       & \multicolumn{3}{c}{position} & \multicolumn{3}{c}{velocity} &    &   & \multicolumn{3}{c}{position} & \multicolumn{3}{c}{velocity} \\ \cmidrule(lr){3-5} \cmidrule(lr){6-8} \cmidrule(lr){11-13} \cmidrule(lr){14-16}
    $K$                   & $w$ [pixel]   &   $x$ [pixel]   &   $y$ [pixel]   &  $r/10^{-2}$ [rad]  &   $v_x$ [pixel/s]    &   $v_y$ [pixel/s]    &   $\omega$ [rad/s]    & $K$                   & $w$ [pixel]   &   $x$ [pixel]   &   $y$ [pixel]   &  $r/10^{-2}$ [rad] &   $v_x$ [pixel/s]   &   $v_y$ [pixel/s]   &   $\omega$ [rad/s] \\ \cmidrule(lr){1-8} \cmidrule(lr){9-16}
    \multirow{5}{*}{300}  &  2    &  1.065  &  2.121  &   0.013            &   3.040    &   3.179    &     0.021   &  \multirow{5}{*}{3000} &  2    &  0.773  &   1.794 &  0.034             &  0.692     &   0.416    &   0.430 \\
                          &  4    &  1.238  &  2.641  &   0.009            &   2.074    &   1.541    &     0.069   &&  4    &  1.753  &   0.967 &  0.023             &  1.635     &   4.334    &   0.130  \\
                          &  8    &  0.623  &  8.191  &   0.014            &   2.680    &   4.174    &     0.081   &&  8    &  0.246  &   0.281 &  0.006             &  0.253     &   0.363    &   0.032   \\
                          & 16    &  1.424  &  3.113  &   0.124            &   1.403    &   1.807    &     0.091   && 16    &  2.635  &   0.201 &  0.011             &  0.208     &   0.919    &   0.206    \\ \cmidrule(lr){2-8}  \cmidrule(lr){10-16}
                          &Ave.   &  1.088  &  4.016  &   0.040            &   2.299    &   2.675    &     0.065   &&Ave.   &  1.352  &   0.811 &  0.019             &  0.697     &   1.508    &   0.199 \\ \cmidrule(lr){1-8} \cmidrule(lr){9-16}
    \multirow{5}{*}{500}  &  2    &  1.090  &  2.878  &   0.013            &   2.859    &   3.173    &     0.018   &\multirow{5}{*}{5000} &  2    &  0.759  &   1.783 &  0.033             &  0.673     &   0.381    &   0.374 \\
                          &  4    &  1.292  &  2.919  &   0.009            &   1.905    &   1.472    &     0.071   & &  4    &  1.728  &   0.962 &  0.023             &  1.525     &   3.522    &   0.132 \\
                          &  8    &  0.646  &  8.919  &   0.013            &   2.512    &   4.056    &     0.079   &&  8    &  0.250  &   0.280 &  0.006             &  0.253     &   0.354    &   0.028 \\
                          &  16   &  1.438  &  3.121  &   0.125            &   1.383    &   1.796    &     0.092    & &  16   &  2.629  &   0.201 &  0.011             &  0.203     &   0.880    &   0.205\\ \cmidrule(lr){2-8} \cmidrule(lr){10-16}
                          &Ave.   &  1.117  &  4.459  &   0.040            &   2.165    &   2.624    &     0.065   & &Ave.   &  1.342  &   0.807 &  0.018             &  0.665     &   1.234    &   0.185  \\ \cmidrule(lr){1-8} \cmidrule(lr){9-16}
    \end{tabular}
    }
\end{table*}   

\begin{table}[tb]
    \centering
    \caption{Average Number of Iteration}
    \label{tab:inference_epoch}
    \begin{tabular}{ccccc}
    \hline
       & \multicolumn{2}{c}{White Background} & \multicolumn{2}{c}{Wood Grain Background}\\ \cmidrule(lr){2-3} \cmidrule(lr){4-5}
    $w$& $K=300$ &$K=500$  &$K=3000$  &$K=5000$ \\ \hline
    2  & 8.65    & 13.66   & 16.57    & 22.66   \\ 
    4  & 8.72    & 19.45   & 20.37    & 27.75    \\
    8  & 11.89   & 20.64   & 18.16    & 27.64   \\ 
    16 & 11.80   & 20.90    & 18.82     & 29.83   \\ \hline
    Ave.& 10.27  & 18.66   & 18.48    & 26.97   \\ \hline
    \end{tabular}
    
\end{table}

\subsection{Experimental Setup}

We evaluated the performance of our algorithm by tracking AR Marker in two environments; one is on the white background and the other is on the wood grain pattern. Since our proposed tracking algorithm is limited to $2$ dimension, we fixed the event camera on the dolly as shown in Fig. \ref{fig:taking_event} so that the distance between the camera and the AR marker is constant. We use optimizer Adam \cite{kingma2014adam} with the recommended hyperparameter settings of: $\beta_1=0.9 $, $\beta_2 = 0.999$, and $\epsilon=10^{-8}$ for training IEG and stochastic gradient descent (SGD) for gradient-based optimization. The learning rate was $0.0001$ for both training and inference and kept constant. We set the threshold of synthetic event firing as $\bar{\delta} = 0.001$ and the optimization threshold as $\bar{\epsilon} = 1.0 \times 10^{-6}$.

\begin{figure}[tb]
    \centering
    \includegraphics[width=0.8\hsize]{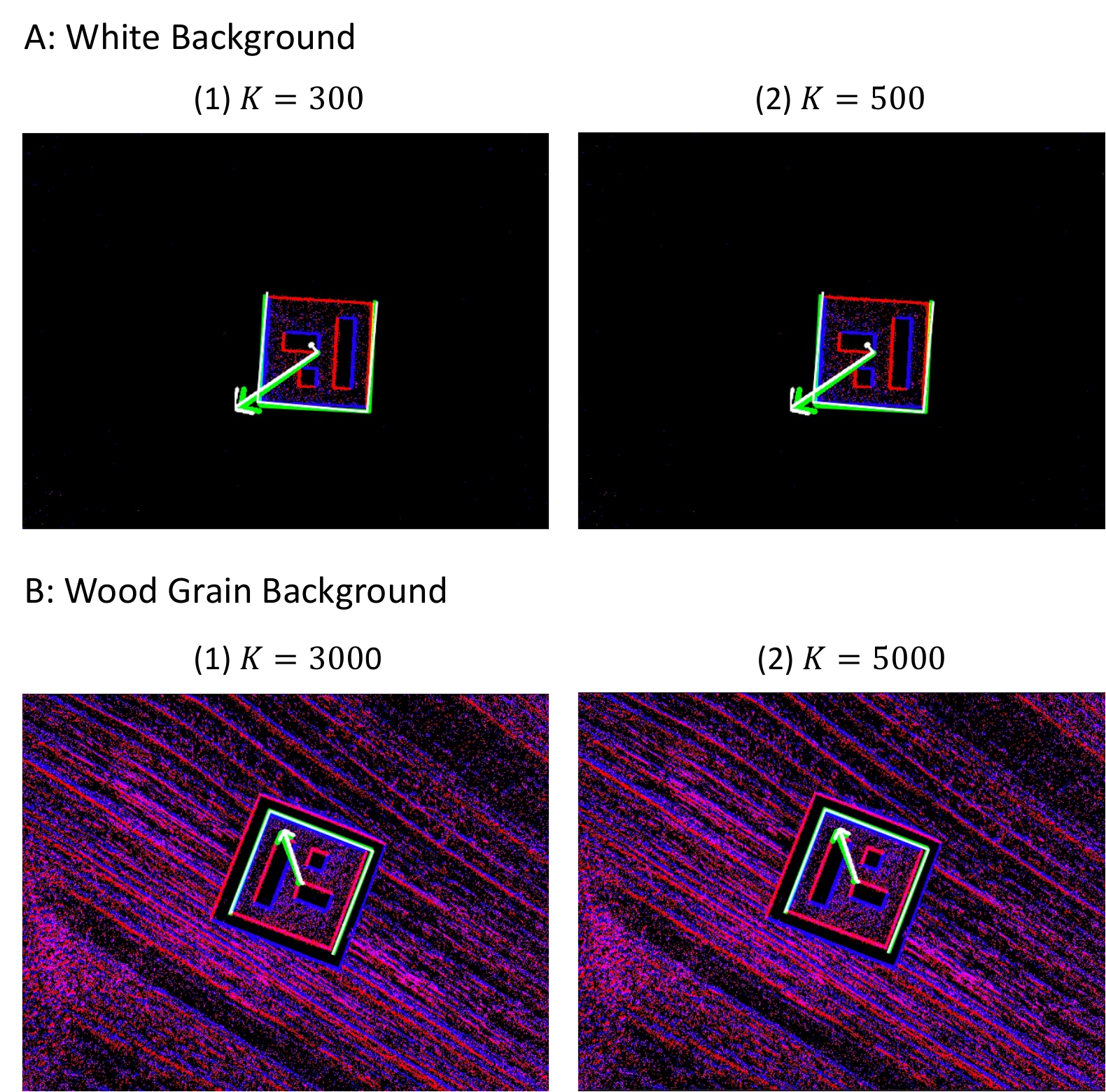}
    \caption{Tracking result visualization of A: on the white background and B: on the wood grain background when $w = 16$. A-(1) is when $K=300$, A-(2) is when $K=500$, B-(1) is when $K=3000$ and B-(2) is when $K=5000$. The white box indicates the predicted position of the AR marker, the white arrows indicate the velocity and rotation speed, The green box indicates the ground truth position of the AR marker and the green arrows indicate the ground truth velocity and rotation speed. Positive events are shown in red and negative events are shown in blue.}
    \label{fig:result_with_gt}
\end{figure}

\paragraph{On the White Background}

For on the white background data, we conduct the experiment with $M = 2\times10^4$ and $K = 300/500$ and evaluated for every $1.5\times10^{4}$ events, i.e. for every five iterations until the $5000$-th iteration for $K = 300$ and for every three iterations until the $3000$-th iteration for $K = 500$.

\paragraph{On the Wood Grain Background}
For on the wood grain background data, we conduct the experiment with $M = 2\times10^5$ and $K = 3000/5000$ and evaluated for every $1.5\times10^{5}$ events, i.e. for every five iterations until the $500$-th iteration for $K = 3000$ and for every three iterations until the $300$-th iteration for $K = 5000$.

\subsection{Tracking Errors}
The position estimation error and velocity estimation error are given by the mean square error for each value and we conduct experiments for $w=2,4,8,16$. The result of the data with the white background the wood grain pattern is shown in Table \ref{tab:quantitative}.

In the case of on white background and on wood grain background, the wood grain background is supposed to be a more difficult problem setting for object tracking, but by comparing the result of on white background and on wood grain background. In this experiment, stable tracking was achieved regardless of the width of the Gaussian blur $w$ and we can see that the tracking performance does not deteriorate even in the case of on wood grain pattern. This shows the possibility that our gradient-based algorithm can work effectively in various situations.

\begin{figure*}[tb]
    \centering
    \includegraphics[width=0.75\hsize]{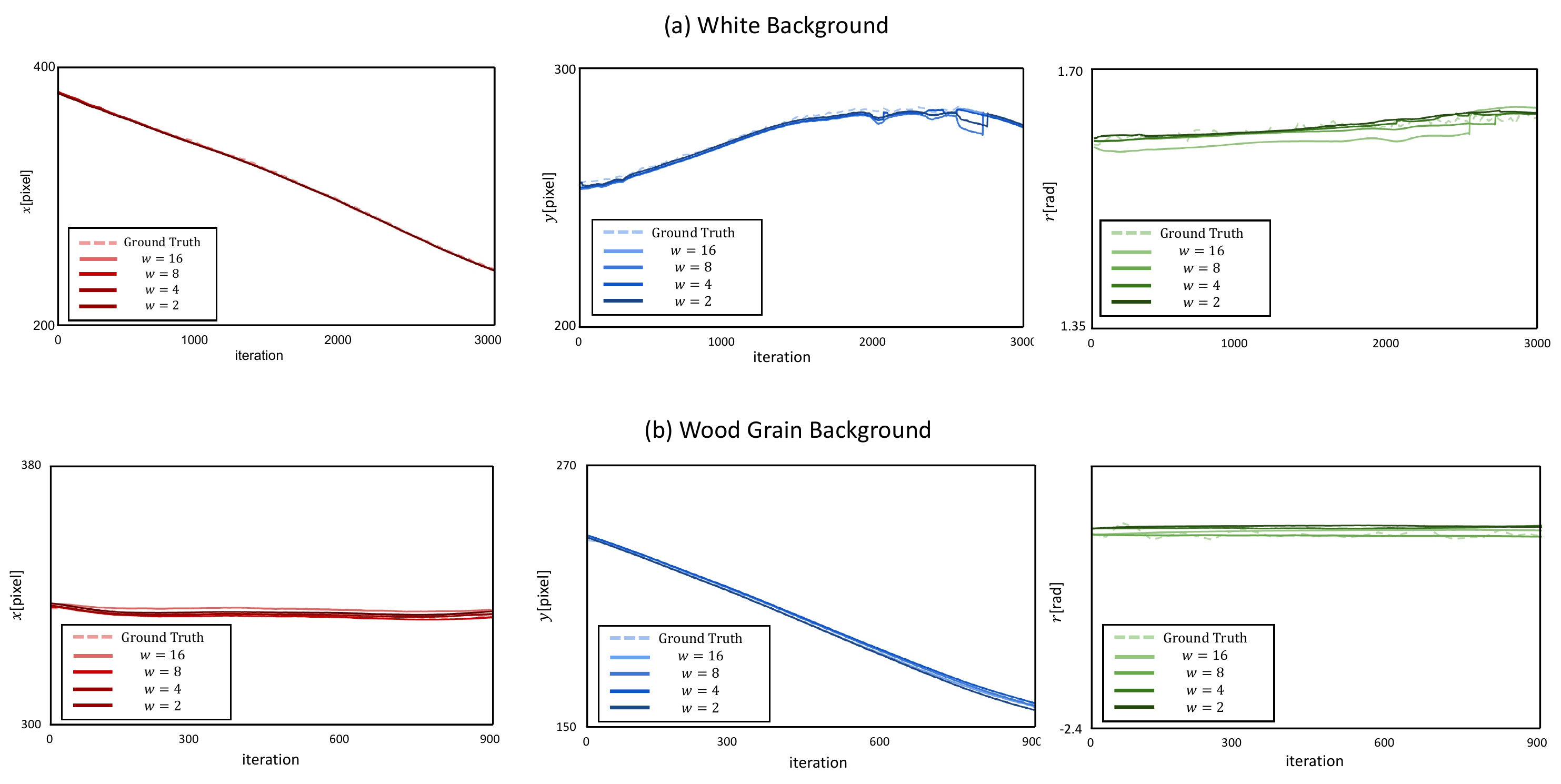}
    \caption{Results of position estimation (solid line) and ground truth (dashed line). (a) shows the result of on a white background, (b) shows that of on a wood grain background and from left to right, these are the results for $x$, $y$, and $r$. These results show that not only is the difference from the ground truth small but also that the misalignment does not increase over time. Also, the tracking performance does not depend on the white background or the wood grain background.}
    \label{fig:transition}
\end{figure*}

\subsection{Number of Iterations}
The average numbers of iterations for $3000$ times of optimization for white background and $300$ times of optimization for wood grain pattern after the second optimization are shown in Table \ref{tab:inference_epoch}. The first optimization required a large number of iterations compared to the second optimization to localize the object, $1301.25$ for the white background and $1684.25$ for the wood grain background. As can be seen from Table \ref{tab:inference_epoch}, the number of optimizations increases with larger $K$, which indicates that the larger the motion of the object, it takes time to convergence.

\subsection{Qualitative Evaluation}
\paragraph{Visualization of tracking results}
Fig. \ref{fig:result_with_gt}-A shows the event of AR marker on white background with tracking result in white box (position) and white arrow (velocity) and ground truth in green box (position) and green arrow (velocity). Fig. \ref{fig:result_with_gt}-B shows the event of AR marker on wood grain background with tracking result and ground truth as in Fig. \ref{fig:result_with_gt}-A. This shows our algorithm can track the movements with small errors for real data regardless of the size of sliding width $K$. In addition to that, as shown in Fig. \ref{fig:result_with_gt}-B, our algorithm gracefully tracks the object precisely despite the considerable amount of events generated by the background.

\paragraph{Robustness over time}
The transition of estimated position $T = (x, y, r)$ (solid line) and ground truth position (dashed line) is shown in Fig. \ref{fig:transition}. (a) shows the result of on the white background and (b) shows that of on the wood grain background. These results show that not only is the difference from the ground truth small but also that the misalignment does not increase over time. Also, the tracking performance did not depend on whether the background was wood grain or white. This indicates that our method is not affected by a large number of events from the background, and can perform gradient optimization and find the position properly and accurately in every iteration.

\section{CONCLUSION}
We proposed a novel concept of motion tracking using implicit expression. Our method estimates an object's position and velocity by gradient-based optimization using our proposed implicit event generator (IEG). Unlike conventional approaches that explicitly generate events, this method generates events implicitly. It makes it possible to handle events sparsely, thereby realizing highly efficient processing, especially suitable for mobile robotics applications. We show that our framework could work in the practical scenario through evaluation using event data capture using a real event-based camera. 

\subsection{Limitation and Future Work}
The purpose of this study is to verify the principle of our novel concept of estimating position and velocity by gradient-based optimization using IEG as shown in Fig. \ref{fig:overview}, and for comparison with existing methods, we need to consider the extension to $6$-DoF, speed-up of inference time, and improvement of Optimizer.

\paragraph{Extention to 6-DoF}
For performance verification, in this paper, we have applied it to a simple $3$-DoF problem and confirmed its operation. In the future, we would like to extend the idea to $6$-DoF, as shown in \cite{bryner2019event,gehrig2018asynchronous,gehrig2020eklt}. The problem with extending this method to $6$-DoF is that the number of variables that need to be updated increases. It is necessary to consider a new scheme for updating each variable appropriately using backpropagation, rather than simply updating each variable as in this proposed method.

\paragraph{Speed up using Look-up table}
In this paper, we have only verified our concept and have not studied the reduction of inference time, which is necessary for practical applications such as AR/VR. To conduct the fast inference, we think it is an effective algorithm that reduces inference time by combining look-up tables and MLP, as proposed by Sekikawa \etal~\cite{sekikawa2020irregularly}.

\paragraph{Optimizer}
In this paper, we used SGD as the optimizer for inference. However, it is necessary to consider optimizers for speeding up and stabilizing inference, such as Momentum \cite{qian1999momentum}, Nesterov's accelerated gradient \cite{nesterov1983method}. In addition, we would like to consider the application of the Gauss-Newton method or the Levenberg-Marquardt method, which use the (approximate) second-order derivatives.

\addtolength{\textheight}{-9cm}   




\bibliographystyle{IEEEtran}
\bibliography{refs}

\end{document}